%% file: main.tex
\PassOptionsToPackage{table}{xcolor}

\documentclass[10pt,twocolumn,letterpaper]{article}

\usepackage[pagenumbers]{iccv} 
\usepackage{booktabs}   
\usepackage{multirow}   


\definecolor{iccvblue}{rgb}{0.21,0.49,0.74}
\usepackage[pagebackref,breaklinks,colorlinks,allcolors=iccvblue]{hyperref}

\title{Less is More: Improving Motion Diffusion Models with Sparse Keyframes}

\author{Jinseok Bae\\
Seoul National University\\
{\tt\small capoo95@snu.ac.kr}
\and
Inwoo Hwang\\
Seoul National University\\
{\tt\small inusu0818@snu.ac.kr}
\and
Young Yoon Lee\\
Roblox\\
{\tt\small ylee@roblox.com}
\and
Ziyu Guo\\
The Chinese University of Hong Kong\\
{\tt\small ziyuguo@link.cuhk.edu.hk}
\and
Joseph Liu\\
Roblox\\
{\tt\small josephliu@roblox.com}
\and
Yizhak Ben-Shabat\\
Roblox\\
{\tt\small ibenshabat@roblox.com}
\and
Young Min Kim\\
Seoul National University\\
{\tt\small youngmin.kim@snu.ac.kr}
\and
Mubbasir Kapadia\\
Roblox\\
{\tt\small mkapadia@roblox.com}
}

\begin{document}

\maketitle


\input{LaTeX/0_abstract}
\input{LaTeX/1_intro}

\input{LaTeX/2_related_works}
\input{LaTeX/3_preliminary}
\input{LaTeX/4_method}
\input{LaTeX/5_experiments}

\input{LaTeX/6_conclusion}

{
    \small
    \bibliographystyle{ieeenat_fullname}
    \bibliography{main}
}

\input{LaTeX_Supplementary/main_supplementary}


\end{document}

%% file: LaTeX/0_abstract.tex
\begin{abstract}
Recent advances in motion diffusion models have led to remarkable progress in diverse motion generation tasks, including text-to-motion synthesis.
However, existing approaches represent motions as dense frame sequences, requiring the model to process redundant or less informative frames.
The processing of dense animation frames imposes significant training complexity, especially when learning intricate distributions of large motion datasets even with modern neural architectures. 
This severely limits the performance of generative motion models for downstream tasks.
Inspired by professional animators who mainly focus on sparse keyframes, we propose a novel diffusion framework explicitly designed around sparse and geometrically meaningful keyframes.
Our method reduces computation by masking non-keyframes and efficiently interpolating missing frames. 
We dynamically refine the keyframe mask during inference to prioritize informative frames in later diffusion steps.
Extensive experiments show that our approach consistently outperforms state-of-the-art methods in text alignment and motion realism, while also effectively maintaining high performance at significantly fewer diffusion steps.
We further validate the robustness of our framework by using it as a generative prior and adapting it to different downstream tasks. 
\end{abstract}

%% file: LaTeX/1_intro.tex
\begin{figure}[t]
    \centering
    \includegraphics[width=\linewidth]{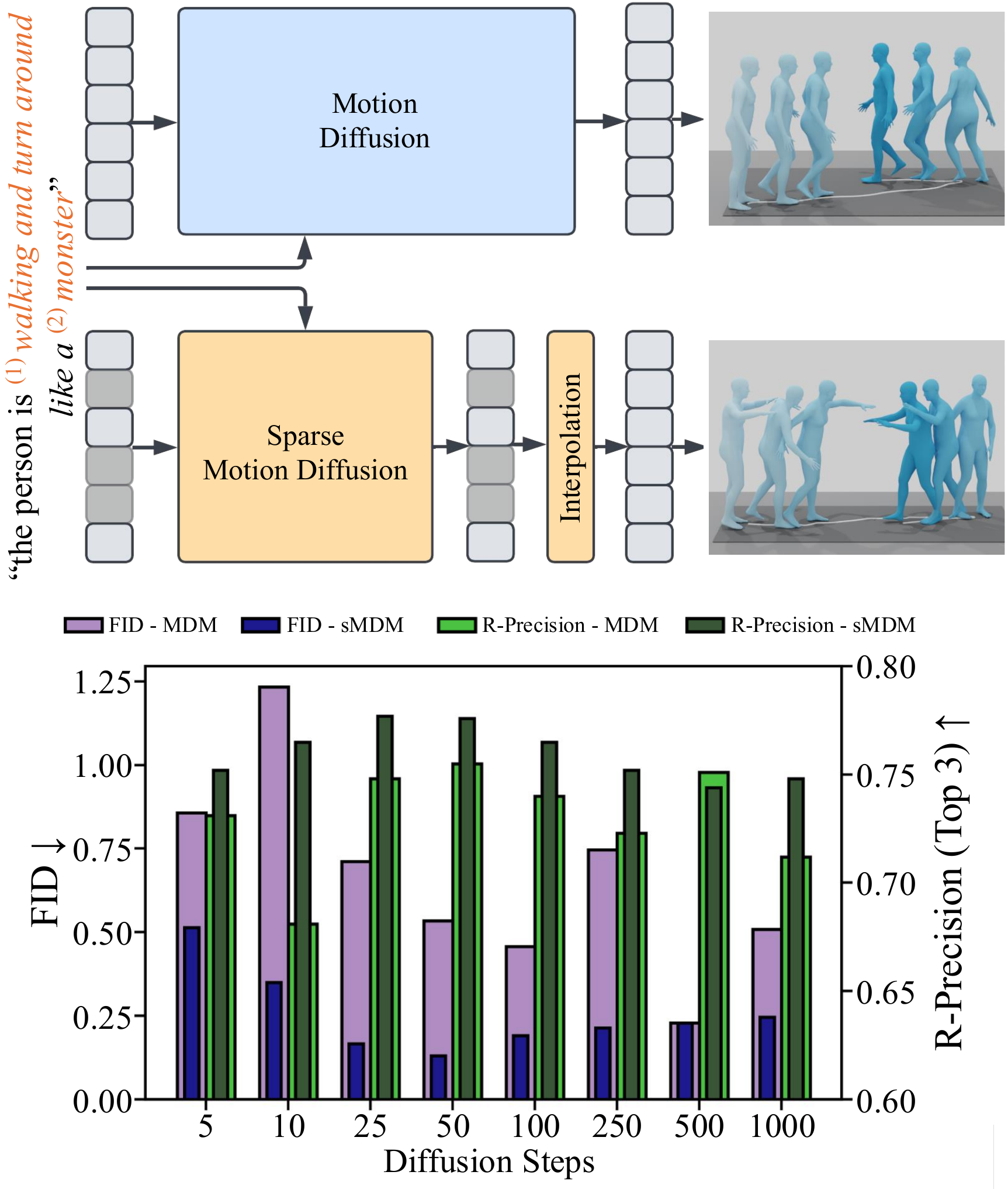}
    \caption{
        We propose a \emph{keyframe-centric} framework for training motion diffusion models.
        Our method, namely \emph{Sparse Motion Diffusion Model (sMDM)}, outperforms baseline \emph{Motion Diffusion Model (MDM)}~\cite{tevet2022human}, achieving more stable and precise motion generation while better capturing text prompts.
    }
    \vspace{-4 mm}
    \label{fig:teaser_fid}
\end{figure}

\section{Introduction}\label{sec:intro}
Human motion generation is a fundamental problem in computer vision and graphics, with applications ranging from animation to robotics. 
Recent advancements have been accelerated by large-scale motion datasets~\cite{mahmood2019amass,harvey2020robust,mahmood2019amass}, enabling diffusion-based models to generate highly realistic motions~\cite{tevet2022human, chen2023executing, dai2024motionlcm}.
Leveraging denoising diffusion processes~\cite{ho2020denoising}, these methods iteratively refine generated motions, achieving impressive realism across various applications, including character animation~\cite{shi2024amdm}, robot control~\cite{serifi2024robot}, and immersive media~\cite{goel2024iterative}.

Existing motion diffusion models typically rely on dense frame sequences, leading to two primary issues:
\noindent\emph{(1) Complexity in Training and Inference.}  
Processing dense frames significantly increases computational overhead.  
This inefficiency arises because self-attention layers, which are often employed in the state-of-the-art architectures, scale quadratically with the number of frames~\cite{vaswani2017attention}.
Moreover, modern datasets often contain vast amounts of complex motion data, making efficient training even more challenging.  
In practice, motion diffusion models frequently degrade at small diffusion step settings, partly due to the burden of learning intricate distributions from such dense input.
Simply reducing frames na\"ively is not a feasible solution, as text-conditioned models require sufficient temporal resolution to maintain alignment with the given prompts.
\noindent\emph{(2) Lack of Controllability.}  
Generating all frames simultaneously makes it difficult to interpret and control the model’s behavior. 
In contrast, professional animators typically define motions through sparse keyframes, clearly delineating essential motion states before interpolating intermediate frames~\cite{goel2024iterative}. 
Dense-frame models, however, treat each frame equally, making it difficult to isolate critical motion points and often compromising the overall quality and realism of generated motions.

Motivated by conventional workflows for animation creation, we propose a \emph{keyframe-centric} approach to address the challenges of dense-frame motion diffusion models.
In our framework, keyframes serve as distinct, informative frames from which the entire motion sequence is effectively represented.
Our model generates motions from keyframes, which aligns better with how humans naturally perceive and create motions.
This is achieved using a straightforward \emph{masking-and-interpolation} strategy within a established transformer architectures. 
During self-attention, insignificant frames are masked out, notably enhancing training efficiency. 
Dense sequences are then reconstructed via ligthweight linear interpolation. 
In order to preserve the high-frequency motion details and ensure high quality motion generation, we constrain the input and output layers using Lipschitz regularization~\cite{liu2022learning} which encourages learning smooth motion manifolds.

Selecting salient keyframes is critical to preserving essential motion content while avoiding unnecessary overhead.
Simple approaches, such as random or uniformly spaced sampling, often fail to adapt to the underlying motion dynamics, resulting in suboptimal learning signals and degraded motion quality.
Instead, we adopt a heuristic-guided keyframe reduction algorithm~\cite{visvalingam1993line} to automatically identify geometrically meaningful frames that minimize reconstruction error.
During training, this algorithm pinpoints frames that capture the most salient motion features, ensuring the model learns an effective denoising prior from a sparse yet informative set of keyframes.
At inference time, we no longer have access to ground-truth keyframe annotations.
To initialize, we employ a uniform mask as a strong prior in the early sampling steps, stabilizing the denoising process.
However, we further refine this mask at later stages through a \emph{dynamic mask update}, adaptively prioritizing the most informative frames. 
This two-stage strategy ensures that the model benefits from both the stabilizing effect of uniform masking in initial denoising and the enhanced motion fidelity afforded by adaptively reallocating keyframe attention in subsequent steps.

We comprehensively evaluate our approach on several motion generation tasks. 
We first examine text-to-motion generation, where precise text alignment and motion quality are essential.  
Our model not only improves these aspects but also maintains robust performance across various diffusion step settings, demonstrating its stability in different configurations.  
Next, we explore long-sequence generation to test its ability as a strong generative prior.  
Generating extended motion sequences requires both expressiveness and consistency, and our model effectively maintains both while faithfully adhering to text commands.  
Finally, we assess its generalizability by applying it to a character control task using an autoregressive motion diffusion model.  
Our approach outperforms baselines, reinforcing its flexibility and effectiveness beyond standard text-to-motion tasks.

%% file: LaTeX/2_related_works.tex
\section{Related Works}

\subsection{Motion Generative Models}

Over the past decade, human motion generation has seen significant advancements through various methodologies. 
These include Variational Autoencoders (VAEs)~\cite{petrovich2021action, rempe2021humor}, Generative Adversarial Networks (GANs)~\cite{tulyakov2018mocogan, cai2018deep}, and autoregressive models~\cite{guo2020action2motion}. 
More recently, diffusion-based frameworks have emerged as robust solutions for modeling rich spatio-temporal dynamics. 
These models utilize an iterative denoising process to refine random noise into high-fidelity motion sequences. 
They have demonstrated strong performance in tasks such as motion in-betweening, text-conditioned motion generation, and multi-person interaction~\cite{tevet2022human, zhang2022motiondiffuse, chen2023executing, yuan2023physdiff, dai2024motionlcm}.

Among diffusion-based methods, the Human Motion Diffusion Model (MDM)~\cite{tevet2022human} stands out as a seminal work. 
MDM popularized the use of denoising diffusion for human motion synthesis. 
Its transformer-based architecture effectively captures the temporal structure of motion, while its conditioning mechanisms allow for flexible text-to-motion generation. 
Subsequent studies have expanded upon MDM by integrating specialized modules for physics-based realism~\cite{yuan2023physdiff, tevet2024closd}, long-horizon synthesis~\cite{yang2023synthesizing, zhuo2024infinidreamer}, and non-human character animation~\cite{kapon2024mas, raab2024single}. 
Recent studies have also shown that pretrained MDM can be extended to various downstream tasks, such as long motion generation~\cite{shafir2023human}, style transfer~\cite{raab2024monkey}, and spatial control scenarios~\cite{liu2024programmable}, in a zero-shot manner. 
Despite these advancements, most existing approaches maintain a frame-dense representation throughout both training and inference. 
This results in high training complexity that does not necessarily translate into improved motion quality.

\subsection{Keyframe-Based Approaches}

Keyframes have been a fundamental concept in filmmaking, defining the starting and ending points of smooth transitions~\cite{parent1988computer}. 
Recent studies have shown that training video generation models using only keyframes can yield significant benefits. 
For example, a previous work on human rendering trains its neural rendering model solely on a sparse set of keyframes extracted from RGBD videos~\cite{pang2021few}. 
This approach effectively captures the essential motion manifold while dramatically reducing redundant computations and memory overhead. 
Similarly, Keyframe-Intermediate model (KEYIN)~\cite{pertsch2019keyin} exploits the subgoal structure inherent in video sequences by focusing on a compact set of informative keyframes. 
This preserves critical motion dynamics and temporal cues that are often diluted in dense frame sequences. 
These findings suggest that sparse keyframe training not only enhances computational efficiency but also enables the model to learn more robust representations by concentrating on the most salient temporal transitions.

In the field of animation, keyframing is a widely used technique, especially for motion editing. 
Previous works have demonstrated that motion diffusion models can effectively perform motion in-betweening~\cite{cohan2024flexible} or language-based iterative editing~\cite{goel2024iterative} when trained with keyframes. 
However, these works are tailored for specific purposes and are not applicable to general-purpose motion synthesis. 
Additionally, they often rely on randomly selected keyframes, which deviates from real-world creation processes.

In contrast, our work systematically identifies geometrically meaningful keyframes through the Visvalingam-Whyatt algorithm~\cite{visvalingam1993line}. 
Building upon the transformer-based diffusion paradigm of MDM, we implement a \textit{masking-and-interpolation} strategy. 
This strategy selectively attends to keyframes while leveraging linear interpolation to reconstruct dense motion sequences. 
As a result, we not only reduce the training complexity but also maintain, or even surpass, the motion quality of state-of-the-art diffusion models. 
Moreover, our dynamic sampling procedure during inference allows the keyframe set to adapt over time. 
This better aligns with real-world animation workflows, where motion is iteratively refined.

%% file: LaTeX/3_preliminary.tex
\begin{figure*}[t]
    \centering
    \includegraphics[width=\textwidth]{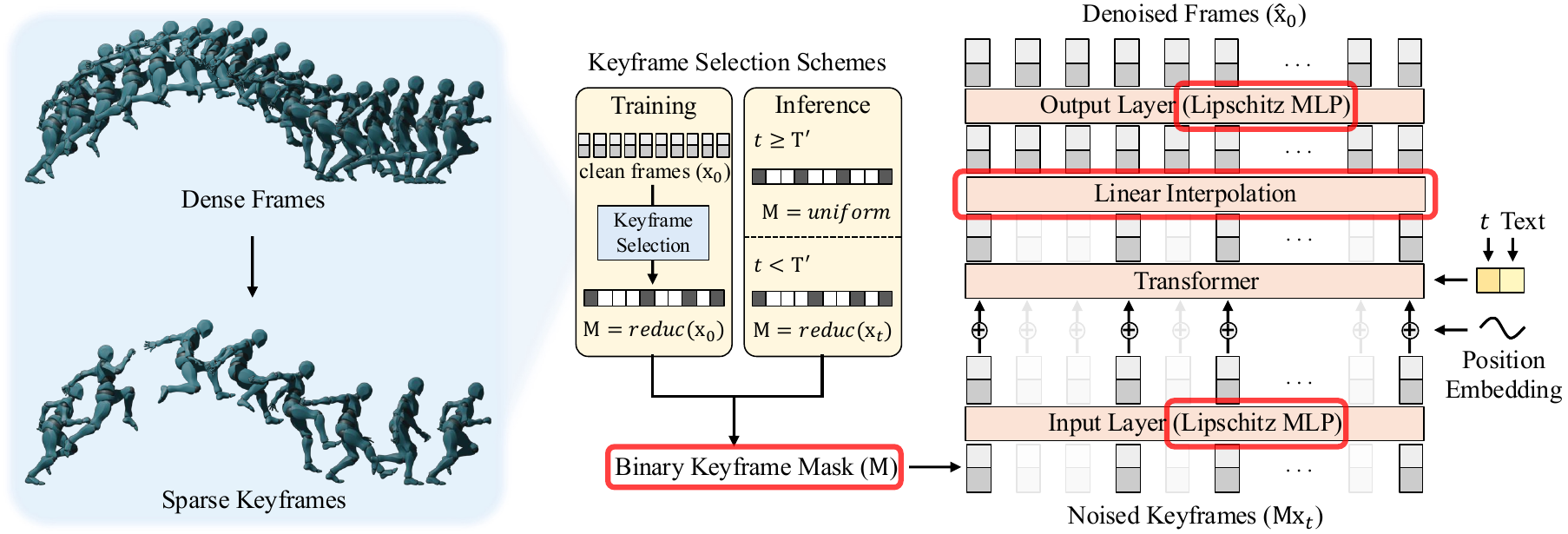}
    \caption{
    Model architectures of Sparse Motion Diffusion Model (sMDM).
    Our sMDM uses a binary keyframe mask $\mathbf{M}$ to exclude non-keyframes from the self-attention layers. During training, $\mathbf{M}$ is derived from the clean input $\mathbf{x}_0$ via keyframe selection~\cite{visvalingam1993line}. At inference, the model starts with a \textit{uniform} keyframe mask at earlier timesteps ($t > T'$), then updates $\mathbf{M}$ by selecting keyframes from $\mathbf{x}_t$ for later timesteps ($t \leq T'$). Finally, sMDM reconstructs the dense motion by linearly interpolating features of the selected keyframes. To ensure smooth interpolation, we replace the input and output linear layers with Lipschitz MLPs~\cite{liu2022learning}. Red boxes indicate the changes from the baseline MDM~\cite{tevet2022human}.
    }
    \label{fig:model_arch}
\end{figure*}

\section{Preliminaries}
\label{sec:prelim}
This section briefly summarizes the diffusion process and denoising network architecture of the seminal motion diffusion model~\cite{tevet2022human}, which serves as the backbone of our approach.

\subsection{Diffusion Process}
\label{sec:prelim_diff_process}

MDM follows the core principles of Denoising Diffusion Probabilistic Models (DDPM)~\cite{ho2020denoising}, which have shown remarkable success in domains such as text-to-image generation.  
In DDPM, the forward diffusion process incrementally adds Gaussian noise to a clean sample over a series of timesteps, transforming it into near-random noise by the final step $T$.  
At each timestep $t$, the noisy sample $\mathbf{x}_t$ is generated from $\mathbf{x}_{t-1}$ as follows:  
\begin{equation}\label{eq:ddpm_forward}
    q(\mathbf{x}_t \mid \mathbf{x}_{t-1}) 
    = \mathcal{N}\bigl(\sqrt{\alpha_t}\,\mathbf{x}_{t-1},\;\alpha_t\,\mathbf{I}\bigr),
\end{equation}
where $\alpha_t$ governs the noise scale at each timestep and is scheduled such that 
$\alpha_1 < \alpha_2 < \cdots < \alpha_T$.  
By taking advantage of Gaussian properties, we can also express $\mathbf{x}_t$ in closed form as a function of the original clean sample $\mathbf{x}_0$.  
When $T$ is sufficiently large, $\mathbf{x}_T$ approaches a standard Gaussian distribution.

The inverse diffusion process, also referred to as sampling, iteratively denoises $\mathbf{x}_T$ back to a clean sample $\mathbf{x}_0$ over $T$ steps.  
Different parameterizations are possible in DDPM, such as predicting $\mathbf{x}_0$, $\mathbf{x}_{t-1}$, or the noise $\boldsymbol{\epsilon}_t$ at each step.  
Following MDM~\cite{tevet2022human}, we adopt the strategy of directly estimating the clean sample $\hat{\mathbf{x}}_0$, which has been shown to be most effective in practice.  
Accordingly, the training objective for the denoising network $f_\theta$ can be written as:  
\begin{equation}\label{eq:ddpm_loss}
    \mathcal{L} \;=\; \bigl\| \mathbf{x}_0 \;-\; \hat{\mathbf{x}}_0 \bigr\|^2,
\end{equation}
where $\hat{\mathbf{x}}_0=f_\theta(x_t, t, c)$ and $c$ is condition such as text.
By minimizing this loss, the model learns to invert the forward noising process, enabling the synthesis of high-fidelity motion from an initial noise vector.

\subsection{Denoising Network}
\label{sec:prelim_model_arch}

The denoising network in MDM~\cite{tevet2022human} is built on a transformer~\cite{vaswani2017attention}, which is well-suited for sequential data like motion.  
To process a motion sequence of length $N$, each frame is represented as a $D$-dimensional vector, forming a 2D matrix $\mathbf{x}_t \in \mathbb{R}^{N \times D}$.  
These noisy input frames are first projected into a higher-dimensional feature space through an input layer.  
Next, the transformer layers operate on the frame-wise feature vectors, incorporating positional embeddings as well as embeddings for text and the current diffusion timestep $t$.  
Through self-attention and feed-forward operations, the network refines these features into intermediate representations.  
Finally, an output layer maps these representations back to the motion space, producing the denoised frames $\hat{\mathbf{x}}_0$.  
Our proposed method modifies this backbone by focusing on a sparse set of keyframes within the self-attention layers, thereby reducing computational overhead while preserving motion fidelity.

%% file: LaTeX/4_method.tex
\section{Sparse Motion Diffusion Model}
\label{sec:method}

We propose sMDM, a sparse keyframe-centric method for training and inference with the motion diffusion models.
In Sec.~\ref{subsec:masking}, we introduce a \emph{masking-and-interpolation} strategy that focuses selected keyframes for computing features while interpolating intermediate frames in feature space.  
In Sec.~\ref{subsec:keyframe-sampling}, we explain how we select keyframes during training and dynamically refine them at inference time.

\subsection{Masking and Interpolation with Keyframes}
\label{subsec:masking}
In a typical motion diffusion model, self-attention involves all frames in the input sequence.  
This leads to high complexity because the calculating attention scales quadratically with sequence length~\cite{vaswani2017attention}.  
Meanwhile, professional animators primarily focus on keyframes that define the motion’s structure~\cite{goel2024iterative}.  
They then generate in-between frames by interpolation or other tools.  
Inspired by this workflow, we modify MDM to attend only to a small set of keyframes.  
By doing so, our model significantly reduces attention operations while retaining important motion cues.

Figure~\ref{fig:model_arch}\, illustrates our framework.  
We create a binary mask \(\mathbf{M}\) that marks each selected keyframe as 1 and non-keyframes as 0.  
During self-attention, only keyframe features are used.  
This reduces attention complexity from \(\mathcal{O}(N^2)\) to roughly \(\mathcal{O}(K^2)\), where \(K \ll N\).  
Here, \(N\) is the total number of frames, and \(K\) is the number of keyframes.  
Afterwards, we linearly interpolate in the \emph{feature space} of keyframes to reconstruct features for the non-keyframes.  
This is computationally efficient as it involves only a few operations per non-keyframe.
Furthermore, as linear interpolation is differentiable, we can still optimize the model through minimizing the prediction error calculated on the dense frames (Eq.~(\ref{eq:ddpm_loss})) without modifications.
From our experiments, we find that models trained using prediction errors computed solely from the input sparse keyframes generate less natural motion compared to those trained with prediction errors computed over the interpolated frames.

Although linear interpolation is fast, real-world motions often have high-frequency details.  
These details might not align well with a purely linear transition in feature space.  
To address this, we replace the basic linear mappings (in both input and output modules) with \emph{Lipschitz MLPs}~\cite{liu2022learning}.
A Lipschitz MLP layer \(g_\theta\) satisfies
\begin{equation}
    \|g_\theta(y_1) - g_\theta(y_2)\|_p \leq \alpha \|y_1 - y_2\|_p,
    \label{eq:lipschitz_continuous}
\end{equation}
where \(\alpha\) is the Lipschitz constant.  
This constraint ensures that small changes in the input space do not lead to disproportionately large changes in the output.  
Thus, the network produces smoother transitions, which reduces artifacts introduced by interpolation.  
We incorporate this idea into the diffusion loss by adding a regularization term \(\lambda \mathcal{L}_{\text{lip}}\):
\begin{equation}
    \mathcal{L} \;=\; \bigl\| \mathbf{x}_0 \;-\; \hat{\mathbf{x}}_0 \bigr\|^2 \;+\; \lambda \mathcal{L}_{\text{lip}},
    \label{eq:our_ddpm_loss}
\end{equation}
where \(\mathbf{x}_0\) is the clean motion, \(\hat{\mathbf{x}}_0\) is the denoised output, and \(\lambda\) is a hyperparameter that balances reconstruction fidelity against smoothness.  
The details of \(\mathcal{L}_{\text{lip}}\) are given in the supplementary material.
Additionally, we employ sinusoidal activation functions~\cite{sitzmann2020implicit} as activations for Lipschitz MLPs.  
Sine-based activations, commonly seen in implicit neural representations, are known to capture fine-grained details and high-frequency variations more effectively than standard activations like ReLU~\cite{sitzmann2020implicit}.  
We empirically find that they are more suitable for modeling intricate aspects of motion, such as sudden direction changes or quick limb movements.

\subsection{Keyframe Selection Schemes}
\label{subsec:keyframe-sampling}
During training, we use the Visvalingam-Whyatt reduction algorithm~\cite{visvalingam1993line} to select frames to label as keyframes.  
This algorithm discards frames with the least geometric importance until it reaches the desired number of keyframes.  
In practice, these keyframes often capture critical motion points, \textit{e.g.}, arm swing peaks or turning points in locomotion.  
We treat the reduction rate as a fixed hyperparameter during training.  
However, to make the model more robust to variations in the number of keyframes, we add a small amount of random noise to the keyframe mask before each training iteration.

At inference time, we do not have access to ground-truth keyframes.  
Hence, we begin with a \emph{uniform} mask: we distribute keyframes evenly throughout the sequence with the reduction rate experienced during the training.  
Although uniform masking does not explicitly favor temporal regions that include rich movements, it provides a stable baseline that ensures coverage across the entire motion span.

We additionally introduce \textit{dynamic mask update}, a technique that further enhances the advantages of keyframes during inference.  
As the diffusion process unfolds, input frames become reliable enough that the keyframe reduction algorithm can detect more meaningful frames.  
Let \(t\) be the diffusion timestep, which decreases from \(T\) to 1.  
When \(t\) drops below \(T' = \gamma \cdot T\), we refine the keyframe set by applying Visvalingam-Whyatt~\cite{visvalingam1993line} to the intermediate frames \(\mathbf{x}_t\).  
This updates the mask to focus attention on the frames most critical at that stage of denoising.  
From experiments, we find \(\gamma = 0.1\) consistently improves motion quality.  
By discarding unimportant frames and selecting more semantically significant ones, the model can better fix remaining errors and generate coherent, realistic motions.  

%% file: LaTeX/5_experiments.tex
\input{Tables/t2m_results}
\input{Tables/dynamic_ablation_results}

\section{Experiments}
In this section, we present how our keyframe approach improves motion quality and text alignment.
We first demonstrate its effectiveness in the core task of text-conditioned motion generation (Sec.~\ref{sec:exp_t2m}).
Next, we further validate its use as a generative prior for long-sequence generation (Sec.~\ref{sec:exp_long_motion}).  
Finally, we exhibit the generalizability of our method by adapting it to an autoregressive motion diffusion model tailored for character control (Sec.~\ref{sec:exp_character_control}).  
Additional training details and visualized results including failure cases, are provided in the supplementary material.

\begin{figure*}[t]
    \centering
    \includegraphics[width=\textwidth]{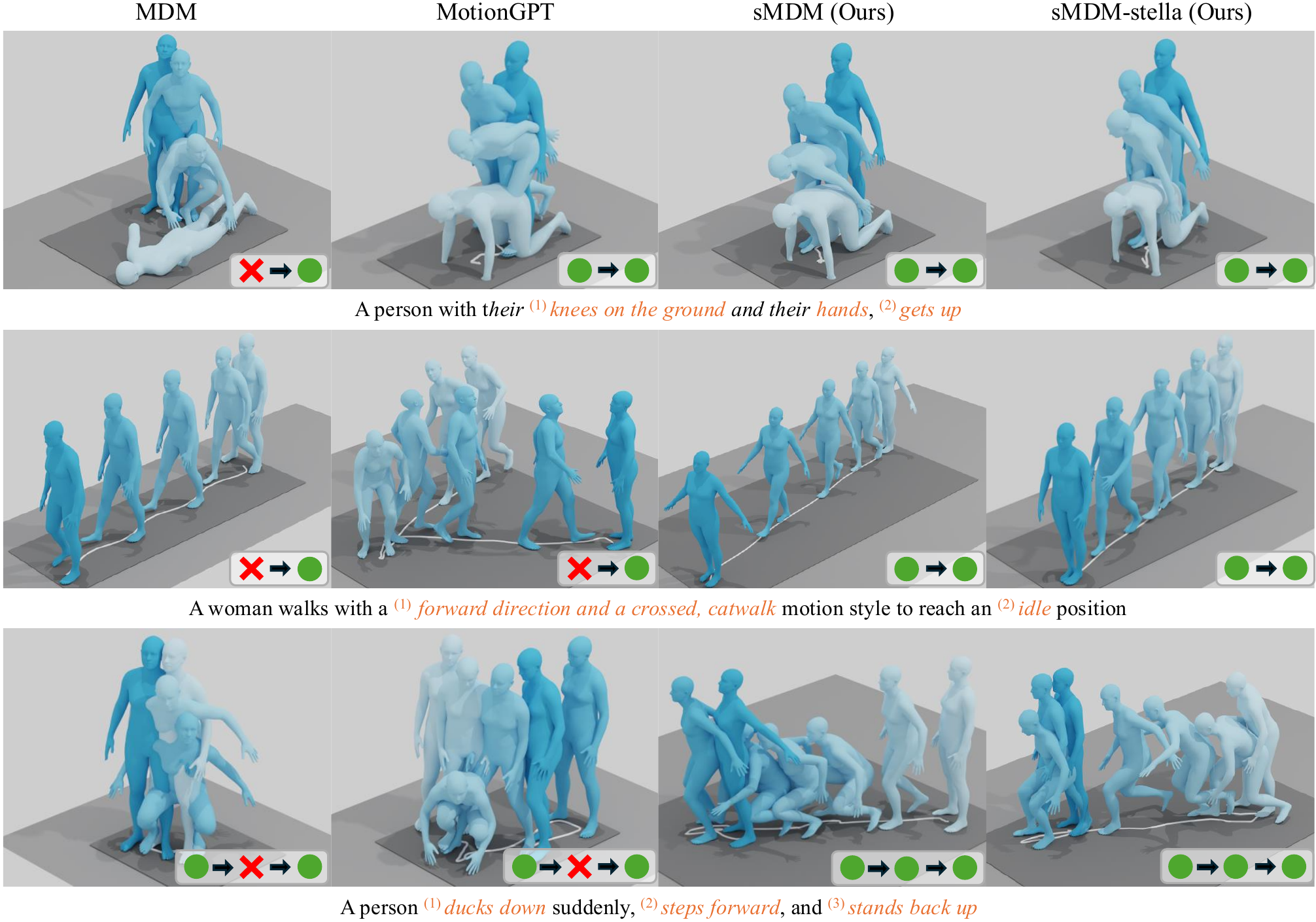}
    \caption{Qualitative evaluations on text-to-motion generation results. We indicate the evaluation results with a \textcolor{Red}{red} cross or \textcolor{ForestGreen}{green} circle, highlighting which parts of the text prompt are missing in the generated motions. Unlike our method, MDM~\cite{tevet2022human} frequently overlooks parts of the input text command. Similarly, although MotionGPT~\cite{jiang2023motiongpt} utilizes an advanced text encoder like sMDM-stella, it struggles to capture fine-grained styles (middle) and contextual details (bottom). In contrast, our models faithfully adhere to the input text commands.}
    \label{fig:t2m_results}
\vspace{-4 mm}
\end{figure*}

\begin{figure*}[t]
    \centering
    \includegraphics[width=\textwidth]{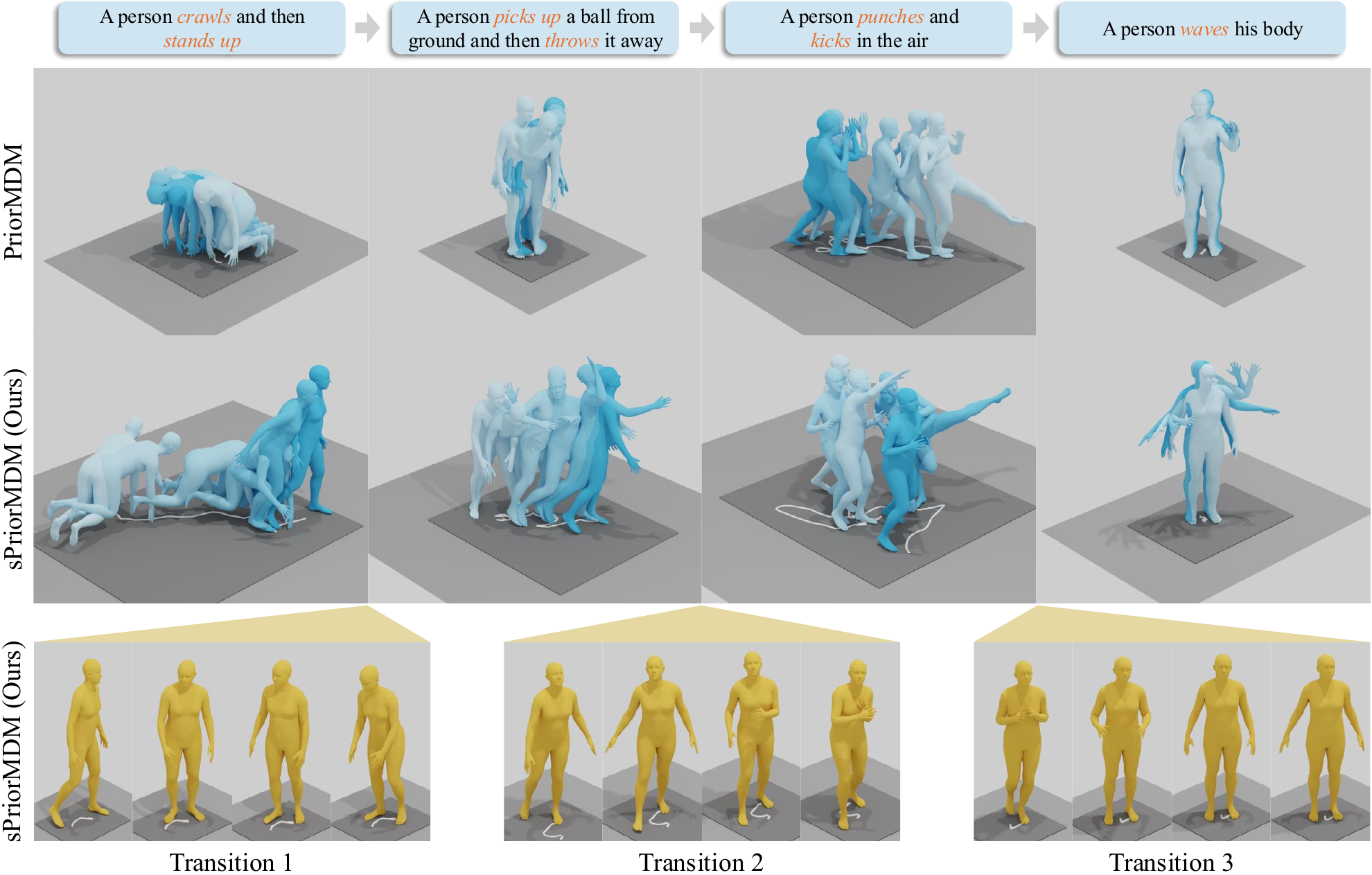}
    \caption{
    Visualization of long-sequence motion generation using the Double Take strategy~\cite{shafir2023human}. 
    Given four sequential text prompts, the pretrained model generates a continuous motion sequence by conditioning on the input prompts.
    To enhance interpretability, we separate each segment in a different frame. 
    PriorMDM and sPriorMDM refer to models using the standard MDM and our sparse MDM (sMDM), respectively.
    Consistent with our findings in regular text-conditioned motion generation (Table~\ref{tab:humanml3d}), our model effectively captures fine-grained motion details described in the prompts, whereas the baseline model frequently fails to fully adhere to the input instructions.
    Furthermore, ours can generate natural transitions that are smoothly aligned with the neighboring motion segments. 
    }
    \label{fig:priormdm_results}
\vspace{-4 mm}
\end{figure*}

\input{LaTeX/5a_t2m}

\input{LaTeX/5b_long_motion}

\input{Tables/priormdm_results}

\input{LaTeX/5c_character_control}

%% file: Tables/t2m_results.tex
\begin{table}[t]
\centering

\renewcommand{\arraystretch}{1.1}
\resizebox{\linewidth}{!}{%
\begin{tabular}{lcccccc}
\toprule
\multirow{2}{*}{Method}
& \multicolumn{3}{c}{R-Precision~$\uparrow$} 
& \multirow{2}{*}{FID$\downarrow$}
& \multirow{2}{*}{\begin{tabular}{c}MM-\\Dist$\downarrow$\end{tabular}}
& \multirow{2}{*}{Div$\rightarrow$} \\
\cmidrule(lr){2-4}
 & Top1 & Top2 & Top3 & & & \\
\midrule
\rowcolor{gray!15}
Real & 0.511 & 0.703 & 0.797 & 0.002 & 2.794 & 9.503 \\ 
\midrule
T2M~\cite{guo2022generating} & 0.457 & 0.639 & 0.740 & 1.067 & 3.340 & 9.188 \\
MotionDiffuse~\cite{zhang2022motiondiffuse} & 0.491 & 0.681 & \underline{0.782} & 0.630 & 3.113 & 9.410 \\
MDM~\cite{tevet2022human} & 0.320 & 0.498 & 0.611 & 0.544 & 5.566 & \underline{9.559} \\
MLD~\cite{chen2023executing} & 0.481 & 0.673 & 0.772 & 0.473 & 3.196 & 9.724 \\
M2DM~\cite{kong2023priority} & \underline{0.497} & \underline{0.682} & 0.763 & 0.352 & 3.134 & 9.724 \\
GMD~\cite{karunratanakul2023guided} & 0.366 & 0.549 & 0.652 & 0.235 & 5.254 & 9.726 \\
CondMDI~\cite{cohan2024flexible} & 0.369 & 0.559 & 0.672 & 0.322 & 4.274 & 9.676 \\
ReMoDiffuse\textsuperscript{\S}~\cite{zhang2023remodiffuse} & \textbf{0.510} & \textbf{0.698} & \textbf{0.795} & \textbf{0.103} & \textbf{2.974} & 9.018\\
\rowcolor{gray!15}
sMDM (Ours) & 0.494 & \underline{0.682} & 0.776 & \underline{0.130} & \underline{3.051} & 9.663 \\
\rowcolor{gray!15}
\;\;\; w random keyframes & 0.493 & 0.677 & 0.777 & 0.249 & 3.086 & 9.602\\
\rowcolor{gray!15}
\;\;\; w/o interpolation & 0.486 & 0.675 & 0.773 & 0.267 & 3.127 & \textbf{9.524} \\
\rowcolor{gray!15}
\;\;\; w/o Lipschitz & 0.474 & 0.665 & 0.768 & 0.329 & 3.149 & 9.384\\
\midrule
MotionGPT\textsuperscript{\textdagger}~\cite{jiang2023motiongpt} & 0.492 & 0.678 & 0.792 & 0.230 & 3.043 & \textbf{9.528} \\
MotionLCM\textsuperscript{\textdagger\ddag}~\cite{dai2024motionlcm} & 0.502 & \underline{0.698} & 0.798 & 0.304 & 3.012 & \underline{9.607} \\
MotionBase\textsuperscript{\textdagger}~\cite{wang2024quo} & \underline{0.519} & - & \underline{0.803} & \underline{0.166} & \underline{2.964} & - \\
\rowcolor{gray!15}
sMDM-stella\textsuperscript{\textdagger} (Ours) & \textbf{0.554} & \textbf{0.740} & \textbf{0.829} & \textbf{0.151} & \textbf{2.740} & 9.797 \\
\bottomrule
\end{tabular}
}
\caption{
Text-to-motion generation results using motion generative models evaluated on HumanML3D dataset~\cite{guo2022generating}.  
The top section presents various motion diffusion models utilizing the same CLIP text encoder~\cite{radford2021learning}, while the bottom section compares other generative models incorporating larger text encoders.  
\textdagger~denotes models that employ an advanced text encoder instead of the standard CLIP~\cite{radford2021learning} text encoder, while \S, \ddag~indicate models that require a retrieval from the database and an additional distillation stage, respectively.  
\textbf{Bold} values indicate the best results, while \underline{underlined} values denote the second-best in each section.  
}

\vspace{-3 mm}
\label{tab:humanml3d}
\end{table}

%% file: Tables/dynamic_ablation_results.tex
\begin{table}[t]
\centering

\renewcommand{\arraystretch}{1.1}
\resizebox{\linewidth}{!}{%
\begin{tabular}{lcccccc}
\toprule
\multirow{2}{*}{Method (Steps)}
& \multicolumn{3}{c}{R-Precision~$\uparrow$} 
& \multirow{2}{*}{FID$\downarrow$}
& \multirow{2}{*}{\begin{tabular}{c}MM-\\Dist$\downarrow$\end{tabular}}
& \multirow{2}{*}{Div$\rightarrow$} \\
\cmidrule(lr){2-4}
 & Top1 & Top2 & Top3 & & & \\
\midrule
\rowcolor{gray!15}
Real & 0.511 & 0.703 & 0.797 & 0.002 & 2.794 & 9.503 \\ 
\midrule
sMDM (1000) & 0.461 & 0.644 & 0.747 & 0.291 & 3.288 & 9.425 \\
\;\;\; w/ dynamic sample & \textcolor{RoyalBlue}{0.466} & \textcolor{RoyalBlue}{0.647} & \textcolor{RoyalBlue}{0.748} & \textcolor{RoyalBlue}{0.246} & \textcolor{RoyalBlue}{3.258} & \textcolor{RoyalBlue}{9.510} \\

\midrule
sMDM (500) & 0.446 & 0.635 & 0.739 & 0.322 & 3.335 & 9.275 \\
\;\;\; w/ dynamic sample & \textcolor{RoyalBlue}{0.451} & \textcolor{RoyalBlue}{0.638} & \textcolor{RoyalBlue}{0.744} & \textcolor{RoyalBlue}{0.229} & \textcolor{RoyalBlue}{3.294} & \textcolor{RoyalBlue}{9.369} \\

\midrule
sMDM (250) & 0.464 & 0.651 & 0.751 & 0.263 & 3.262 & 9.396 \\
\;\;\; w/ dynamic sample & \textcolor{RoyalBlue}{0.465} & \textcolor{RoyalBlue}{0.653} & \textcolor{RoyalBlue}{0.752} & \textcolor{RoyalBlue}{0.214} & \textcolor{RoyalBlue}{3.248} & \textcolor{RoyalBlue}{9.447} \\

\midrule
sMDM (100) & 0.479 & 0.664 & 0.763 & 0.230 & 3.193 & 9.411 \\
\;\;\; w/ dynamic sample & \textcolor{RoyalBlue}{0.480} & \textcolor{RoyalBlue}{0.666} & \textcolor{RoyalBlue}{0.765} & \textcolor{RoyalBlue}{0.190} & \textcolor{RoyalBlue}{3.124} & \textcolor{RoyalBlue}{9.459} \\

\midrule
sMDM (50) & 0.494 & 0.682 & 0.776 & 0.130 & 3.051 & 9.663 \\
\;\;\; w/ dynamic sample & \textcolor{RoyalBlue}{0.496} & \textcolor{RoyalBlue}{0.683} & 0.776 & \textcolor{BrickRed}{0.134} & \textcolor{RoyalBlue}{3.050} & \textcolor{BrickRed}{9.690}\\

\midrule
sMDM (25) & 0.499 & 0.681 & 0.777 & 0.167 & 3.080 & 9.629 \\
\;\;\; w/ dynamic sample & \textcolor{RoyalBlue}{0.500} & \textcolor{BrickRed}{0.680} & \textcolor{RoyalBlue}{0.778} & 0.167 & \textcolor{BrickRed}{3.081} & \textcolor{BrickRed}{9.645} \\

\midrule
sMDM (10) & 0.486 & 0.670 & 0.765 & 0.349 & 3.121 & 9.861 \\
\;\;\; w/ dynamic sample & 0.486 & \textcolor{BrickRed}{0.669} & 0.765 & \textcolor{BrickRed}{0.385} & \textcolor{BrickRed}{3.131} & \textcolor{BrickRed}{9.874} \\

\bottomrule
\end{tabular}
}
\caption{Ablation studies on \textit{dynamic mask update} (Sec.~\ref{subsec:keyframe-sampling}). This strategy is particularly effective in terms of motion quality (FID) for the models trained with relatively large 
diffusion steps.}
\vspace{-5 mm}
\label{tab:ablation_dyna}
\end{table}

%% file: LaTeX/5a_t2m.tex
\subsection{Text-Conditioned Motion Generation}\label{sec:exp_t2m}

We train and evaluate our model using HumanML3D~\cite{guo2022generating}, a large text-paired motion dataset containing about 30 hours of diverse motion from AMASS~\cite{mahmood2019amass}.  
To assess generation quality, we adopt standard metrics from prior works~\cite{guo2020action2motion, guo2022generating}: R-Precision, FID, multi-modal distance (MM-Dist.), and Diversity.  
Generally, R-Precision and MM-Dist. are highly related to text alignment, while FID captures the overall realism of the generated motions.

We train two variants of our method, denoted \emph{sMDM} and \emph{sMDM-stella}.  
Like many motion diffusion models~\cite{chen2023executing, zhang2022motiondiffuse, tevet2022human}, sMDM uses CLIP ViT-B/32 (63M) as its text encoder.  
In contrast, sMDM-stella employs the larger Stella-1.5B~\cite{zhang2024jasper} to explore the compatibility of our approach with more advanced text encoders.  
Following recent methods based on MDM (e.g., MOMO~\cite{raab2024monkey} and CLoSD~\cite{tevet2024closd}), we use a transformer decoder as the core denoising network unless otherwise specified.  
Our approach, however, is not restricted to this specific architecture (see supplementary).
Also, we use 80\% of reduction rate for running keyframe reduction algorithm. 

We compare our approach against various motion generation models on the HumanML3D test set in Table~\ref{tab:humanml3d}.  
In the top section, sMDM outperforms the original MDM~\cite{tevet2022human} by a large margin across most metrics, and its FID is comparable to the retrieval-based ReMoDiffuse~\cite{zhang2023remodiffuse}. 
As shown in Figure~\ref{fig:teaser_fid}, sMDM also sustains superior performance at diverse diffusion step settings compared to the non-keyframe baseline (MDM).  
Notably, it produces high-quality motions even at extremely low diffusion steps, a finding further corroborated by the experiments with 10-step model in Sec.~\ref{sec:exp_character_control}.  
This indicates that training with sparse keyframes can be a promising approach, boosting model performance in reflecting text commands without sacrificing quality.

In the bottom section of Table~\ref{tab:humanml3d}, sMDM-stella outperforms baselines with advanced text encoders, achieving strong text alignment alongside high motion quality.  
For reference, these baselines include MotionGPT~\cite{jiang2023motiongpt} (T5-Base, 220M), MotionLCM~\cite{dai2024motionlcm} (T5-Large, 770M), and MotionBase~\cite{wang2024quo} (Llama-2, 13B).  
Although prior work~\cite{wang2024quo} shows that larger text encoders do not always yield higher-quality motions, our method scales effectively to Stella-1.5B, maintaining both low FID and strong text alignment.  
Moreover, despite retaining the original MDM architecture, we achieve results on par with state-of-the-art models that adopt more complex networks.  
This confirms that focusing on sparse keyframes can guide diffusion-based motion generation effectively, requiring no major architectural overhauls.
Qualitative examples in Figure~\ref{fig:t2m_results} illustrate how our models capture subtle styles or multiple actions, unlike other methods that often overlook parts of the text prompt.

We further examine our design via ablation in Table~\ref{tab:humanml3d}.  
Specifically, we compare: (i) a model that randomly selects keyframes (w random keyframes), (ii) a model that omits interpolation and only computes the loss among the keyframes (w/o interpolation), and (iii) a model without Lipschitz layers (w/o Lipschitz).  
Although some ablated versions produce competitive R-Precision or Diversity scores, they fall short on FID and multi-modal distances, suggesting reduced motion quality.  
This indicates that each component of our model design (interpolation, Lipschitz) and training (keyframe reduction) is essential for training a diffusion model with sparse keyframes.
We also evaluate the impact of dynamic mask update (Sec.~\ref{subsec:masking}) in Table~\ref{tab:ablation_dyna}, finding it particularly beneficial for models using a large number of diffusion steps.  
This aligns with techniques like spatial guidance in OmniControl~\cite{xie2024omnicontrol}, which highlight that later diffusion stages with small noise scale benefit most from additional guidance.  
Our findings confirm that selectively refining keyframes in the final steps allows the model to correct errors more precisely, resulting in smoother and more realistic motions.

%% file: LaTeX/5b_long_motion.tex
\subsection{Long-Sequence Generation}\label{sec:exp_long_motion}

We further demonstrate that sMDM can serve as a generative prior in a downstream task.  
Here, we adopt DoubleTake~\cite{shafir2023human}, which uses a pre-trained MDM or sMDM to generate multiple short motion segments, then revises their overlaps to ensure coherent transitions.
To distinguish between the two pre-trained priors (both with 50 diffusion steps and a transformer encoder), we denote them as \emph{PriorMDM} and \emph{sPriorMDM}.
We set the transition length to 1\,s and use a 0.25\,s margin for blending each segment.

Table~\ref{tab:doubletake} reports quantitative results on these long generated motions, splitting the evaluation into a “Motion” section for individual segments and a “Transition” section for overlapping regions.  
Consistent with our findings from text-to-motion (Sec.~\ref{sec:exp_t2m}), sPriorMDM yields higher-quality segments that better reflect text prompts.  
However, it underperforms in the transition parts, showing lower scores on all metrics.
To investigate this discrepancy, we introduce a new metric, the \emph{End-Effector Speed} (EES), which quantifies the extent of movement in generated motions.  
We define EES as the average end-effector speed (\(\text{cm}/\text{frame}\)), where a higher score corresponds to more dynamic motions.  
Our analysis shows that sPriorMDM generates more expressive and movement-rich segments.  
As illustrated in Figure~\ref{fig:priormdm_results}, sPriorMDM tends to produce active motions in response to commands such as crawling, kicking, or dancing, whereas PriorMDM often generates relatively static movements.  
We attribute this difference to the keyframe-centric training of sMDM, which enables the model to learn realistic motion dynamics from a sparse set of frames.  
Consequently, transitions in sPriorMDM involve dynamic motion prefixes and suffixes, leading to less common motion patterns in the original dataset. 
Although this results in a lower FID score for transitions, qualitative evaluations confirm that sPriorMDM produces smooth and natural transitions between motion segments.

%% file: Tables/priormdm_results.tex
\begin{table}[t]
\centering
\renewcommand{\arraystretch}{1.1}
\resizebox{\linewidth}{!}{%
\begin{tabular}{lccccc|cc}
\toprule
\multirow{2}{*}{Method}
& \multicolumn{5}{c}{Motion} 
& \multicolumn{2}{c}{Transition} \\ 
\cmidrule(lr){2-8}
 & R-Pr~$\uparrow$ & FID$\downarrow$ & Dist$\downarrow$ & Div$\rightarrow$ & EES $\rightarrow$ & FID$\downarrow$ & EES $\rightarrow$ \\
\midrule
\rowcolor{gray!15}
Real & 0.800 & 0.002 & 2.971 & 9.510 & 2.209 & 0.002 & 2.837 \\ 
\midrule
PriorMDM & 0.622 & 0.971 & 5.552 & 10.274 & 1.856 & \textbf{1.125} & \textbf{2.468} \\
sPriorMDM & \textbf{0.660} & \textbf{0.624} & \textbf{5.213} & \textbf{10.115} & \textbf{2.338} & 1.705 & 3.644 \\
\bottomrule
\end{tabular}
}
\caption{Evaluation on Double Take~\cite{shafir2023human} results generated from the model pretrained on HumanML3D dataset~\cite{guo2022generating}. 
We denote results from the experiment using a pretrained standard MDM as PriorMDM, while those using a pretrained sMDM as sPriorMDM. 
Both pretrained models are trained with a 50-step diffusion setting.}
\vspace{-4 mm}
\label{tab:doubletake}
\end{table}

%% file: LaTeX/5c_character_control.tex
\subsection{Real-Time Character Controller}\label{sec:exp_character_control}
We further validate our approach by applying it to \emph{Diffusion Planner (DiP)}, an autoregressive motion diffusion model designed for character control~\cite{tevet2024closd}.  
Unlike standard motion generators, DiP predicts 2 seconds of future motion from 1 second of past trajectories.  
It takes both text commands and spatial control signals as input, allowing motion sequences of arbitrary length with dynamically updated prompts.  

DiP uses a transformer decoder as its denoising network and DistilBERT~\cite{sanh2019distilbert} as the text encoder.  
To support real-time control, it is trained with only 10 diffusion steps.  
It also supports \emph{target conditioning}, where control trajectories from various joints can be provided on the fly.  
This feature makes DiP a versatile model for character control.  
For clarity, we refer to the model without target conditioning as \emph{DiP} and the variant with it as \emph{DiP-T}.  
When training our approach (\emph{sDiP} and \emph{sDiP-T}), keyframe selection is applied only to future frames, while past trajectories remain dense.  

Table~\ref{tab:dip} presents text-to-motion results on the HumanML3D dataset~\cite{guo2022generating}.  
Both sDiP and sDiP-T show improvements in R-Precision, FID, multi-modal distance, and Diversity.  
This is consistent with the previous experiments, where training with sparse keyframes improves both the text alignment and motion realism.  
We also test different keyframe reduction rates (60\%, 70\%, 80\%).  
For sDiP, an 80\% reduction achieves the best text alignment, while 60\% yields the lowest FID.  
For sDiP-T, 70\% provides the strongest text alignment, while 60\% again results in the lowest FID.  
These results suggest that the reduction rate plays a key role in balancing alignment and motion quality.
\input{Tables/dip_results}

%% file: Tables/dip_results.tex
\begin{table}[t]
\centering
\renewcommand{\arraystretch}{1.1}
\resizebox{\linewidth}{!}{%
\begin{tabular}{lcccccc}
\toprule
\multirow{2}{*}{Method}
& \multicolumn{3}{c}{R-Precision~$\uparrow$} 
& \multirow{2}{*}{FID$\downarrow$}
& \multirow{2}{*}{\begin{tabular}{c}MM-\\Dist$\downarrow$\end{tabular}}
& \multirow{2}{*}{Div$\rightarrow$} \\
\cmidrule(lr){2-4}
 & Top1 & Top2 & Top3 & & & \\
\midrule
\rowcolor{gray!15}
Real & 0.511 & 0.703 & 0.797 & 0.002 & 2.794 & 9.503 \\ 
\midrule
DiP & 0.457 & 0.670 & 0.781 & 0.214 & 3.187 & 9.409 \\
sDiP (Reduc. 60\%) & \underline{0.473} & \underline{0.682} & \underline{0.792} & \textbf{0.175} & \underline{3.115} & \textbf{9.468} \\
sDiP (Reduc. 70\%) & 0.469 & 0.677 & 0.789 & \underline{0.186} & 3.130 & \underline{9.419} \\
sDiP (Reduc. 80\%) & \textbf{0.479} & \textbf{0.687} & \textbf{0.797} & 0.199 & \textbf{3.106} & 9.390 \\
\midrule
DiP-T & 0.451 & 0.661 & 0. 771 & 0.243 & 3.218 & 9.312 \\
sDiP-T (Reduc. 60\%) & 0.463 & 0.670 & 0.779 & \textbf{0.198} & 3.180 & \textbf{9.443} \\
sDiP-T (Reduc. 70\%) & \textbf{0.470} & \textbf{0.681} & \textbf{0.791} & \underline{0.213} & \textbf{3.141} & 9.362 \\
sDiP-T (Reduc. 80\%) & \underline{0.467} & \underline{0.678} & \underline{0.787} & 0.218 & \underline{3.163} & \underline{9.371} \\
\bottomrule
\end{tabular}
}
\caption{Text-to-motion results using the Diffusion Planner (DiP) model~\cite{tevet2024closd} on HumanML3D dataset~\cite{guo2022generating}. 
We distinguish between DiP and DiP-T, where DiP-T represents models trained with target conditioning technique for dynamic change of control joint sets, while DiP denotes models trained without it. 
Additionally, sDiP and sDiP-T refer to models trained using our approach, with different keyframe reduction rates specified in parentheses.}
\vspace{-4 mm}
\label{tab:dip}
\end{table}

%% file: LaTeX/6_conclusion.tex
\section{Conclusion}
We present a novel framework for sparse keyframe-based motion diffusion with three main contributions.  
First, we introduce a masking strategy that removes non-keyframes from the network, thereby enforcing model to focus on a sparse set of keyframes.  
We also employ a lightweight linear interpolation module to recover dense frames without complex in-betweening steps. 
This interpolation is further enhanced with Lipschitz MLP layers, providing smoother feature-space transitions and improving motion quality.  
Second, we propose keyframe selection schemes for both training and inference.  
In place of random keyframe choice, we use a keyframe reduction algorithm to provide meaningful frames during training.  
Additionally, our dynamic mask update strategy refines the keyframe mask during inference, further boosting generation quality.  
Finally, extensive experiments on text-to-motion generation and other downstream tasks show that our approach consistently improves text alignment and motion quality, and also serves as a robust generative prior.

Despite these results, our evaluations mainly rely on Transformer-based motion diffusion models.  
Future work may explore extending our approach to UNet-based architectures~\cite{cohan2024flexible} or other generative frameworks~\cite{guo2024momask, jiang2023motiongpt}.  
Since our approach is agnostic to the specific algorithm for the keyframe selection, one interesting direction could be evaluating alternative selection methods with different choice of features.  
Lastly, examining sparse-frame approaches in motion editing and other related tasks presents another promising avenue for future work.

%% file: LaTeX_Supplementary/main_supplementary.tex
\clearpage
\setcounter{page}{1}
\setcounter{section}{0}
\setcounter{equation}{4}
\setcounter{figure}{4}
\setcounter{table}{4}
\def\thesection{\Alph{section}}

\maketitlesupplementary

\input{LaTeX_Supplementary/A_method_details}
\input{LaTeX_Supplementary/B_Implementation_details}

\input{LaTeX_Supplementary/C_additional_results}

%% file: LaTeX_Supplementary/A_method_details.tex
\section{Method Details}
In this section, we provide additional explanations on the methodology introduced in Sec. 4.
Specifically, we describe details about employed Lipschitz MLPs~\cite{liu2022learning} (Sec.~\ref{subsec_supp:lipschitz}) and Visvalingam-Whyatt algorithm~\cite{visvalingam1993line} for keyframe reduction algorithm(Sec.~\ref{subsec_supp:reduction}), respectively.

\subsection{Lipschitz Regularization}\label{subsec_supp:lipschitz}
Lipschitz MLP~\cite{liu2022learning} regularizes a multilayer perceptron (MLP) by bounding the Lipschitz constant $c$ (Eq. (3)). 
By constraining the network's Lipschitz constant, we enforce the network to learn smoother mapping between input and output feature space.
Lipschitz constant $c$ is calculated as the product of norms of the linear layers' weight matrices: $\prod_i{\|\mathbf{W}_i\|_p}$.
Since this bound depends solely on trainable parameters $\mathbf{W}_i$, this regularization method efficiently avoids gradient vanishing problem.
Authors suggest using $\infty$-norm can enhance computational efficiency.

Additionally, we follow original implementation which employs weight normalization technique for further encourage robustness of the network.
Specifically, the authors conduct row-wise normalization where $k$-th row of matrix $\mathbf{W}_i$ is normalized through 
\begin{equation}
    \hat{\mathbf{W}}_{i,k}= \mathbf{W}_{i,k}\cdot\min(1, \frac{\text{softplus}(\|\mathbf{W}_i\|_p)}{\|\mathbf{W}_{i,k}\|_p}),
\end{equation}
where the function $\text{softplus}(y)$ is defined as $\log(1+e^{y})$.
As a result, $i$-th layer of Lipschitz MLP is defined as 
\begin{equation} 
    g_{\theta,i}(x) = \sigma(\hat{\mathbf{W}}_i x + b_i). 
\end{equation} 
The overall Lipschitz regulrization loss $\mathcal{L}_{\text{lip}}$ of the network is then computed as the product of norms, 
\begin{equation}
    \mathcal{L}_{\text{lip}} = \prod_i{\text{softplus}(\|\mathbf{W}_i\|_p)}.
\end{equation}
We recommend to see the further descriptions in the original paper~\cite{liu2022learning}.

\subsection{Keyframe Reduction Algorithm}\label{subsec_supp:reduction}
Although our method is agnostic to specific keyframe reduction algorithms, we select the Visvalingam-Whyatt algorithm~\cite{visvalingam1993line} as our primary method.
The Visvalingam-Whyatt algorithm is a line simplification method that reduces the number of vertices in a polyline while preserving its overall shape. 
It works by iteratively removing the vertex with the smallest \emph{effective area}, defined as the area of the triangle formed by that vertex and its adjacent vertices. 
For a given vertex $v_i$ with neighbors $v_{i-1}$ and $v_{i+1}$ (coordinates $(x_{i-1}, y_{i-1})$, $(x_i, y_i)$, and $(x_{i+1}, y_{i+1})$), the effective area $A_i$ is computed as 
\begin{equation}
A_i = \frac{1}{2}\left|\det\begin{pmatrix}
x_{i-1} & y_{i-1} & 1 \\
x_i     & y_i     & 1 \\
x_{i+1} & y_{i+1} & 1
\end{pmatrix}\right|.
\end{equation}
which equals the area of triangle $\triangle (v_{i-1}, v_i, v_{i+1})$. 
After running this algorithm until the end, we obtain a priority list of indices, where the higher priority means the latter the point is removed.
By cropping certain length from the start, we can get most reliable points that can effectively represent the input geometry.
Here, the reduction rate determines the length to crop, \textit{i.e.} the number of keyframes.
We find 60\% to 80\% of reduction rate often extracts plausible sets of keyframes for motions with 20 Frames per second (FPS).
However, we note that the optimal values can be different depending on the dataset. 

To apply the Visvalingam-Whyatt algorithm to keyframe selection, we first define a suitable feature for each frame. 
Through empirical observation, we find that using joint positions in the root coordinate space yields effective results. 
Specifically, for the SMPL-based HumanML3D dataset~\cite{guo2022generating}, we compute local positions $\mathbf{p}_j \in \mathbb{R}^3$ for all $J=23$ joints.
Additionally, we append an additional dimension for the frame index, which leads extracting evenly distributed keyframes along temporal axis.
Consequently, each frame is represented as a 64-dimensional vector, and we run the algorithm directly in this 64-dimensional space to identify keyframes.

%% file: LaTeX_Supplementary/B_Implementation_details.tex
\section{Implementation Details}
For reproducibility, we additionally provide detailed explanations on the implementation for Sparse Motion Diffusion Model (sMDM, Sec.~\ref{subsec:b_smdm}) and Sparse Diffusion Planner (sDiP, Sec.~\ref{subsec:b_sdip}), respectively.

\subsection{Sparse Motion Diffusion Model (sMDM)}\label{subsec:b_smdm}

We leverage the Transformer-based architecture~\cite{vaswani2017attention} from the baseline MDM~\cite{tevet2022human}.
This implementation employs Transformer layers from PyTorch~\cite{paszke2019pytorch}, and provides options between Transformer encoder and decoder for denoising network.
Following the original implementation, we use 8 layers of Transformer layers, while employing 4 heads for multi-head attention structure.
Each transformer layer introduces Dropout layers~\cite{srivastava2014dropout} with 0.1 of dropout probability.
Also, we allocate 512 dimension for the intermediate features space, \textit{i.e.} using 512-dimensional vector for latent representation.
For additional robustness, we use customized self-attention layer, where key feature are quantized through finite scalar quantization (FSQ)~\cite{mentzer2023finite}.
Unlike to regular vector quantization~\cite{van2017neural}, this scalar quantization does not employ additional loss terms.
Lastly, we exclude Lipschitz regularization as the model without it produces better result.

\subsection{Sparse Diffusion Planner (sDiP)}\label{subsec:b_sdip}
We also follow the official codebase of Diffusion Planner (DiP).
As DiP largely borrows model structure from MDM~\cite{tevet2022human}, we use the same hyperparameters in Sec.~\ref{subsec:b_smdm} for architectural design.
Unlike regular MDM, DiP aims to generate motions in a real-time controller scenario, which demands fast inference time.
Following original implementation, we train this model with 10 diffusion steps, which employing 20 frames (1 sec.) for the past trajectory while 40 frames (2 sec.) for future trajectory (denoising target).
Also, for the \textit{target-conditioning} example, we employ a linear layer with a Sigmoid Linear Unit (SiLU) activation to encode multiple conditioning joint trajectories.
As demonstrated in Table 2, we exclude \textit{dynamic mask update} during the inference, as this strategy is not much effective with the model with the small diffusion step.

%% file: LaTeX_Supplementary/C_additional_results.tex
\section{Additional Results}
\subsection{Results on Different Transformer Layers}
We evaluate our approach's generalizability across different Transformer architectures. 
Although our method is architecture-agnostic, we specifically validate performance improvements on Transformer encoder layers. 
To directly measure our method's impact, we retrain the baseline MDM from scratch rather than using the pre-trained model. Results are shown in Table~\ref{tab:supp_trans_enc}.

\begin{table}[h]
\centering
\renewcommand{\arraystretch}{1.1}
\resizebox{\linewidth}{!}{%
\begin{tabular}{lcccccc}
\toprule
\multirow{2}{*}{Method}
& \multicolumn{3}{c}{R-Precision~$\uparrow$} 
& \multirow{2}{*}{FID$\downarrow$}
& \multirow{2}{*}{\begin{tabular}{c}MM-\\Dist$\downarrow$\end{tabular}}
& \multirow{2}{*}{Div$\rightarrow$} \\
\cmidrule(lr){2-4}
 & Top1 & Top2 & Top3 & & & \\
\midrule
\rowcolor{gray!15}
Real & 0.511 & 0.703 & 0.797 & 0.002 & 2.794 & 9.503 \\ 
\midrule
MDM (enc) & 0.463 & 0.656 & 0.755 & 0.495 & 3.206 & 10.050 \\
sMDM (enc) & \textbf{0.503} & \textbf{0.697} & \textbf{0.793} & \textbf{0.284} & \textbf{3.007} & \textbf{9.955} \\
\bottomrule
\end{tabular}
}
\caption{
    Text-to-motion results of MDM~\cite{tevet2022human} and sMDM, implemented with Transformer encoder, on HumanML3D dataset~\cite{guo2022generating}. 
    Both models are trained with 50 diffusion steps. 
    We use the same evaluation metrics in Table 1.
}
\label{tab:supp_trans_enc}
\end{table}
As shown in Table~\ref{tab:supp_trans_enc}, our method consistently outperforms the baseline, confirming that our approach effectively generalizes across different Transformer architectures.

We further validate the generality of our method by applying it to MotionDiffuse\cite{zhang2022motiondiffuse}, namely \emph{Sparse MotionDiffuse (sMotionDiffuse)}.
To ensure fairness, we directly modified the original MotionDiffuse codebase and retrained the model from scratch. 
Similar to sMDM, we apply an 80\% reduction rate for keyframe selection, keeping all other settings unchanged. 
During implementation, we corrected an error in the original diffusion loss term, where invalid frames were not properly masked.
Table~\ref{tab:supp_motiondiffuse} shows quantitative improvements after integrating our sparse keyframe approach.

\begin{table}[h]
\centering
\renewcommand{\arraystretch}{1.1}
\resizebox{\linewidth}{!}{%
\begin{tabular}{lcccccc}
\toprule
\multirow{2}{*}{Method}
& \multicolumn{3}{c}{R-Precision~$\uparrow$} 
& \multirow{2}{*}{FID$\downarrow$}
& \multirow{2}{*}{\begin{tabular}{c}MM-\\Dist$\downarrow$\end{tabular}}
& \multirow{2}{*}{Div$\rightarrow$} \\
\cmidrule(lr){2-4}
 & Top1 & Top2 & Top3 & & & \\
\midrule
\rowcolor{gray!15}
Real & 0.511 & 0.703 & 0.797 & 0.002 & 2.794 & 9.503 \\ 
\midrule
MotionDiffuse & 0.459 & 0.648 & 0.753 & 1.096 & 3.262 & 9.073 \\
sMotionDiffuse & \textbf{0.483} & \textbf{0.674} & \textbf{0.776} & \textbf{0.698} & \textbf{3.110} & \textbf{9.375} \\
\bottomrule
\end{tabular}
}
\caption{
    Text-to-motion results of MotionDiffuse and sMotionDiffuse, implemented with Transformer encoder, on HumanML3D dataset~\cite{guo2022generating}. 
}
\label{tab:supp_motiondiffuse}
\end{table}
Results indicate consistent improvements across all evaluation metrics, demonstrating that our approach is broadly applicable to Transformer-based motion diffusion frameworks, independent of the specific architectural details.

\subsection{Results on U-Net based Architecture}
We also present preliminary results to assess the generalization of our approach to motion diffusion models with U-Net-based architectures~\cite{ronneberger2015unet}. 

Although initial motion diffusion models, such as MDM~\cite{tevet2022human} and subsequent works, primarily utilized Transformer-based architectures, recent studies have explored U-Net alternatives. 
Notably, Guided Motion Diffusion (GMD)~\cite{karunratanakul2023guided} showed that U-Net architectures are particularly effective for motion representations involving absolute root coordinates. 
Specifically, GMD highlights the advantages of using U-Net when editing motion data with the conditioning trajectories in a global coordintates.
Inspired by GMD, CondMDI~\cite{cohan2024flexible} modify the suggested model with imputation technique in order to conduct flexible motion in-betweening.

Since CondMDI demonstrates the state-of-the-art performance with U-Net structure, we choose CondMDI as the baseline to test our approach.
We summarize three components where CondMDI mainly differs from MDM:
\emph{(1) Architecture of Denoising Network.} Although both architectures employ attention layers, U-Net in CondMDI can not exclusively use sparse keyframes during the calculation. This is because U-Net heavily relies on 1D convolutional layers along temporal axis, which assumes dense sequence as input. \emph{(2) Data Representation.} CondMDI uses modified representation of HumanML3D dataset~\cite{guo2022generating}. It uses absolute positions rather than localized velocities for root representation. \emph{(3) Imputation.} As CondMDI presents motion diffusion model tailored to motion in-betweening task, it explicitly employs imputation process during the training. In contrast to MDM which uses noised motions $\mathbf{x}_t$ as input, CondMDI imputates noised frames with the clean input $\mathbf{x}_0$, resulting 
\begin{equation}
    \hat{\mathbf{x}}_t=\mathbf{M}\cdot \mathbf{x}_0 + (1 - \mathbf{M})\cdot \mathbf{x}_t,
    \label{eq:condmdi}
\end{equation}
where $\mathbf{M}$ corresponds to the keyrame mask.
Different from our approach, the authors of CondMDI propose random selections to create a keyframe mask during the training.

As U-Net with 1D convolution cannot fundamentally accept sparse frames as input, our sparse version, namely \emph{sCondMDI}, simply replaces uninformative keyframes with the learnable embeddings.
Also, we employ similar geometric losses~\cite{tevet2022human} that are originally introduced to prevent foot skating or jerk.
For keyframe selection, we use keyframes selected from Visvalingam-Whyatt algorithm~\cite{visvalingam1993line} with dynamic reduction rate in a range of [90\%, 95\%]. 
We evaluate the motion in-betweening, which measures generation quality given 5 conditioning keyframes, in Table~\ref{tab:supp_condmdi}.

\begin{table}[h]
\centering
\renewcommand{\arraystretch}{1.1}
\resizebox{\linewidth}{!}{%
\begin{tabular}{lcccccc}
\toprule
\multirow{2}{*}{Method}
& \multirow{2}{*}{R-Pre~$\uparrow$} 
& \multirow{2}{*}{FID$\downarrow$}
& \multirow{2}{*}{\begin{tabular}{c}MM-\\Dist$\downarrow$\end{tabular}}
& \multirow{2}{*}{Div$\rightarrow$}
& \multirow{2}{*}{\begin{tabular}{c}Keyframe\\Error$\downarrow$\end{tabular}}
& \multirow{2}{*}{\begin{tabular}{c}Skating\\Ratio$\downarrow$\end{tabular}}
\\
 & & & & & & \\
\midrule
\rowcolor{gray!15}
Real & 0.797 & 0.002 & 2.794 & 9.503 & - & - \\ 
\midrule
CondMDI & \textbf{0.669} & \textbf{0.153} & \underline{5.127} & \textbf{9.457} & \textbf{0.081} & \textbf{0.067}\\
\;\;\; w/o geometric & \textbf{0.669} & \underline{0.322} & 5.157 & 9.031 & \underline{0.096} & 0.098 \\
sCondMDI & 0.667 & 0.551 & \textbf{5.059} & \underline{9.075} & 0.218 & \underline{0.082} \\
\bottomrule
\end{tabular}
}
\caption{
    Motion in-betweening results of CondMDI and sCondMDI. For evaluation, we sample keyframes from test split of HumanML3D dataset~\cite{guo2022generating} using Visvalingam-Whyatt~\cite{visvalingam1993line} algorithm. 
}
\label{tab:supp_condmdi}
\end{table}

Along with the evaluation metrics for generation (R-Precision (Top 3), FID, Multi-Modal Distance, Diversity), we additionally measure the keyframe errors and skating ratio, following the original work.
As demonstrated in the table, our approach exhibits degraded performance compared to the baseline.
We attribute this that our preliminary modification cannot focus on the sparse keyframes.
Future work could explore further architectural modifications to more naturally accommodate sparsity.